\documentclass{article}
\usepackage{nips13submit_e,times}
\usepackage{hyperref}
 \usepackage{amsmath} 
  \usepackage{amssymb} 
\usepackage{natbib}
 \usepackage{bbm}
\usepackage{url}
\usepackage{cases}	
\usepackage{graphicx}
\usepackage{fullpage}
\usepackage{wrapfig}
\usepackage{graphicx}
\usepackage{caption}
\usepackage{subcaption}

\title{Graph Neural Networks and Boolean Satisfiability}
\author{
  Benedikt B\"unz\\
  Department of Computer Science\\
  Stanford University\\
  Stanford, CA 94305 \\
  \texttt{buenz@cs.stanford.edu} \\
   \And
  Matthew Lamm\thanks{The authors contributed equally to this article}
 \\
  Department of Linguistics\\
  Stanford University\\
  Stanford, CA 94305 \\
  \texttt{mlamm@stanford.edu} \\
}

\nipsfinalcopy 

\begin{document}
\maketitle
\begin{abstract}
In this paper we explore whether or not deep neural architectures can learn to classify Boolean satisfiability (SAT). We devote considerable time to discussing the theoretical properties of SAT. Then, we define a graph representation for Boolean formulas in conjunctive normal form, and train neural classifiers over general graph structures called Graph Neural Networks, or GNNs, to recognize features of satisfiability. To the best of our knowledge this has never been tried before. Our preliminary findings are potentially profound. In a weakly-supervised setting, that is, without problem specific feature engineering, Graph Neural Networks can learn features of satisfiability.

\end{abstract}

\section{Introduction}

The Boolean satisfiability problem, or SAT, asks whether there exists a satisfying assignment to the variables of a propositional formula. If such an assignment exists, we say the problem is SAT. If it does not, we say it is UNSAT. It is assumed without loss of generality that formulas are given in conjunctive normal form, or CNF. A formula $\phi$ in CNF is written as a conjunction of $M$ clauses $\omega_{i}$, each of which is a disjunction of literals, $x_{j}$ or $\neg x_{j}$. For example $$(x_{1} \vee \neg x_{2} \vee x_{4}) \wedge ( x_{2} \vee x_{3} ) \wedge ( \neg x_{3} \vee x_{4} ) $$
is a CNF formula.
It has been shown that SAT is NP-complete \citep{Cook:1971:CTP:800157.805047}; this implies that even the hardest problems in NP can be expressed as a SAT problem. On the other hand many SAT problems turn out to be easy in practice. Modern 
SAT solvers exist that can solve extremely large instances of SAT in a matter of milliseconds \citep{Selman95localsearch}. This disparity motivates the search for properties that make SAT instances difficult.

 SAT is a self-reducible problem. Given an oracle determining whether a problem instance is satisfiable or not, one can find a satisfying assignment in time 
linear in the number of variables. This has motivated recent work by \cite{Devlin08b.:satisfiability} examining the performance of a host of machine-learning classifiers for
 satisfiability. In that paper, the authors employ a variety of manually-designed features, many of which encode graph-like properties of CNF formulas such as the average number
 of unique clauses a variable appears in.

Elsewhere, neural networks have shown great promise in reasoning about some subclasses of graphs, such as the tree-like structures of natural language syntax \citep{SocherEtAl2012:MVRNN}. Despite a degree of opacity to what is in fact being “learned” by these models, they are remarkable in their ability to obtain state-of-the art performance in a problem 
space that is traditionally thought to require a great deal of expert knowledge and feature engineering. While the aptness of neural networks for studying Boolean semantics does not fall out of these findings, we maintain that there is an important connection between natural language meaning on the one hand, and formal-logical meaning on the other.

At first pass, we have found that the imposition of tree-like structure onto CNF formulas, in a sense forcing upon them an ad-hoc natural-language syntax, is an unsound approach. Building on these results, and the aforementioned finding that manually-designed graph features can help other kinds of satisfiability classifiers, we define a representation of CNF formulas as a graphs. We then train Graph Neural Networks, or GNNs, to classify problem instances as satisfiable or unsatisfiable \citep{scarselli2009graph}.

Our results are intriguing. GNNs \textit{are} capable of classifying satisfiability. We are still in the process of exploring how formal problem difficulty affects graph representations of CNF formulas, and how exactly this relates to the accuracy of GNN classifiers on these graphs.

In section 2 we briefly review both related work and preliminary experiments. In section 3 we discuss some theoretical properties of Boolean satisfiability, emphasizing implications for training learners on weakly-supervised graphs to discover properties of satisfiability.  In section 4 we review GNNs. In section 5 we describe our graph representation of CNF formulas. In section 6 we present experimental results, and conclude in section 7.

\section{Related Work and preliminary exploration}
\subsection{Circuit solvers}

The idea to use neural architectures to solve combinatorial optimization problems initially gained traction during what might be termed the ``first wave" of neural network research. \cite{hopfield} famously developed a neural architecture that could solve some instances of the traveling salesman problem. In another instance \citet{Johnson1989435} used a neural architecture to encode 3-SAT.

The general scheme of earlier approaches was to define a circuit over which an objective function attains globally optimum values only at satisfying assignments in the Boolean polytope. While these architectures allow for the reinterpretation of analog problems in digital form, in sufficiently complex cases the objective is riddled with local optima that make gradient-based optimization difficult. Moreover, they do not “learn” from data in any way, and in this sense do not exploit the power of neural networks suggested by more recent research. 

See Appendix A for results on our own implementation of a Johnson-style network.

\subsection{Deep learning, NLP, and Boolean logic}

Neural network classifiers have recently achieved high performance on natural language semantics tasks such as sentiment analysis \citep{SocherEtAl2012:MVRNN}. Propositional logic is a different kind of language, 
but one with a semantics no less: The denotation of a Boolean expression is typically thought to be its truth value given an assignment to its variables. Also like natural language, this denotation is the result
of a compositional operation on the truth values of the variables the formula contains.

Recursive Neural Networks have been designed to leverage the tree-like syntactic structures of natural language to capture complex aspects of meaning composition. This is an important point at which the latent structure of a CNF formula diverges from that of natural language. Logical conjunctions and disjunctions are fully commutative and there is no natural interpretation of CNF syntax as being tree-like.

There is a more philosophical distinction to be made about semantic classification here. In natural language there are multiple \textit{kinds of meaning}, of which sentiment is but one example. The difficulty of classifying satisfiability arises in part from the fact that it is not a question of a particular kind of meaning, but rather at a Boolean formula's \textit{capacity to mean}. The task for a learner is not akin to classifying sentiment, or even to finding a valid interpretation. The learner is charged with determining whether or not some valid interpretation, of indeterminate form, exists at all. Satisfiability classification is in this sense a problem of higher-order.

In Appendix B we demonstrate our finding that Recursive Neural Networks over tree-like structures are ineffective at learning about satisfiability.

\section{Theoretical Properties of SAT}
The difficulty of SAT problems exhibits a hard phase-shift phenomenon that has persistently puzzled theorists of computability. In k-SAT problems where the number of atoms per CNF clause is fixed to be exactly k, one observes a drastic increase in the percentage of unsatisfiable problems after a very specific clause-to-literal ratio. In 3-SAT, for example, the phase shift control ratio $\alpha_{cr} \approx 4.3$ \citep{Saitta:2011:PTM:2024601,mlSAT}. 

It is further understood that for problems that fall far enough to the left of the phase shift, the relative Hamming distance among solutions is low. There is another threshold before $\alpha_{cr}$, after which point the solution space is progressively broken up into exponentially many clusters. At the phase shift, then, solutions to formulas that are satisfiable are scattered sparingly in an exponentially large Boolean hypercube, and are difficult find. After $\alpha_{cr}$, an altogether different phenomenon emerges. For the rare satisfiable formulas that exist after the phase-shift, problems are discovered to have a \textit{backbone}. The \textit{backbone} is a set of variables, each of which takes on the same value for every satisfying assignment to a given formula. \citep{Saitta:2011:PTM:2024601}

The phase shift and related phenomena likely have important implications for a learner of features of satisfiability and unsatisfiability. In an unsatisfiable instance some unresolvable conflict arises out of the complex interplay of multiple Boolean variables distributed across multiple conjoined clauses, resulting in a logical contradiction. Given this fact, it already seems a daunting task to develop a neural architecture capable of discovering these features on its own given a set of problem-label pairs.

There are other unanswered questions. Do unsatisfiable formulas occuring after the phase shift, where satisfiabile problems have backbones, look very different from the unsatisfiable problems to the left of the phase shift? Are there different \textit{kinds of} unsatisfiability, each better highlighted by different representations of the relations that inhere among variables and clauses in CNF formulas? These are questions that loom large as we design experiments and assess results.

\section{Graph Neural Networks}

Graph Neural Networks, or GNNs, denote a class of neural networks that implement functions of the form $\tau(G, n) \in \mathbbm{R}^{m}$ which map a graph $G$ and one of its nodes into an $m$-dimensional Euclidean space. \cite{scarselli2009computational} show that GNNs approximate any functions on graphs that satisfy \textit{preservation of unfolding equivalence}. Informally, this means that GNNs fail to produce distinct outputs only when input graphs exhibit certain highly particular symmetries. By implication GNNs are capable of counting node degree, second-order node degree, and detecting cliques of a given size in a set of graphs \citep{scarselli2009graph}.

Given a graph $G = (N, E)$, where $N$ is a set of nodes and $E$ is the set of edges between them, a GNN employs two functions. $h_{w}$ is a parametric function describing the relationships among nodes in a graph. $g_{w}$ models the connection between output labels and the relationships described by $h_{w}$. More specifically the \textit{state} $x_{n}$ of node $n \in G$ is computed as $$x_{n} = \sum_{u\in ne[n]}h_{w} ( l_{n}, l_{n,u}, x_{u} , l_{u} )$$
where $l_{n}$ is the label of node $n$, $l_{n,u}$ is the label of the edge between nodes $n$ and its neighbor $u$, and so forth for $u$. Note the transformation defines a system of equations over all nodes, where the state $x_{n}$ of a node is a function of the states of $x_{u}$ of its neighbors. It is precisely this functional form that allows GNNs to extend beyond the capacity of FNNs and RNNs, because it can be defined for undirected and cyclic graphs. In line with Banach's fixed-point theorem, it is assumed that $h$ is a contraction map with respect to node state so that this system of equations has a stable state.

In Linear GNNs, transition functions are modeled as in graph autocorrelative models:
						$$h_{w} ( l_{n}, l_{n,u}, x_{u} , l_{u} ) = A_{n,u} x_{u} + b_{n}$$
In Non-Linear GNNs, transition functions are instead modeled as multi-layered feed-forward neural networks (FNNs). In this instance, additional penalty terms are required to ensure that the resulting function is still a contraction map.

In graph-level classification, as opposed to node-level classification, the learning set is given by $$\mathcal{L} = \{(G_{i}, n_{i,j}, t_{i,j}) | G_{i} = (N_{i}, E_{i}) \in \mathcal{G}, n_{i,j}\in N_{i}, t_{i,j} \in \mathbbm{R}^{m}, 1\leq i \leq p, q_{i} = j = 1 \}$$
where $n_{i,j}$ is a special output node containing classification or regression targets. For node-level applications $q_{i} \geq 1$. Learning proceeds with a gradient descent method, and gradients are computed  using backpropagation for graph networks. For details, see \citep{scarselli2009graph}.

\section{CNF formulas as graphs}
As previously discussed, the satisfiability of a Boolean formula is a matter of whether or not the formula encodes some (potentially very complex) conflict among the variables over which it is defined. It seems true that if a weakly-supervised algorithm is to learn to recognize these patterns of conflict it must provided at the very least with relational information among variables in a formula. For example, whether a variable is negated in a given formula or whether or not two variables appear in a clause together.

That inter-variable and inter-clausal dependencies define a graph is borne out by the findings of \cite{Devlin08b.:satisfiability} for whom manually-designed graph-like features were found to help classification of satisfiability. This view is additionally motivated by negative results presented in Appendix B, to the effect that neural networks for natural language sentential semantics fail to classify satisfiability. We suspect this is in part due to the fact that those models are designed to operate over structures in which linear ordering is important. As aforementioned, CNFs are commutative, and so dependencies among clauses are not a function of distance in any way.

\begin{figure}
\center
	\includegraphics[scale=.2]{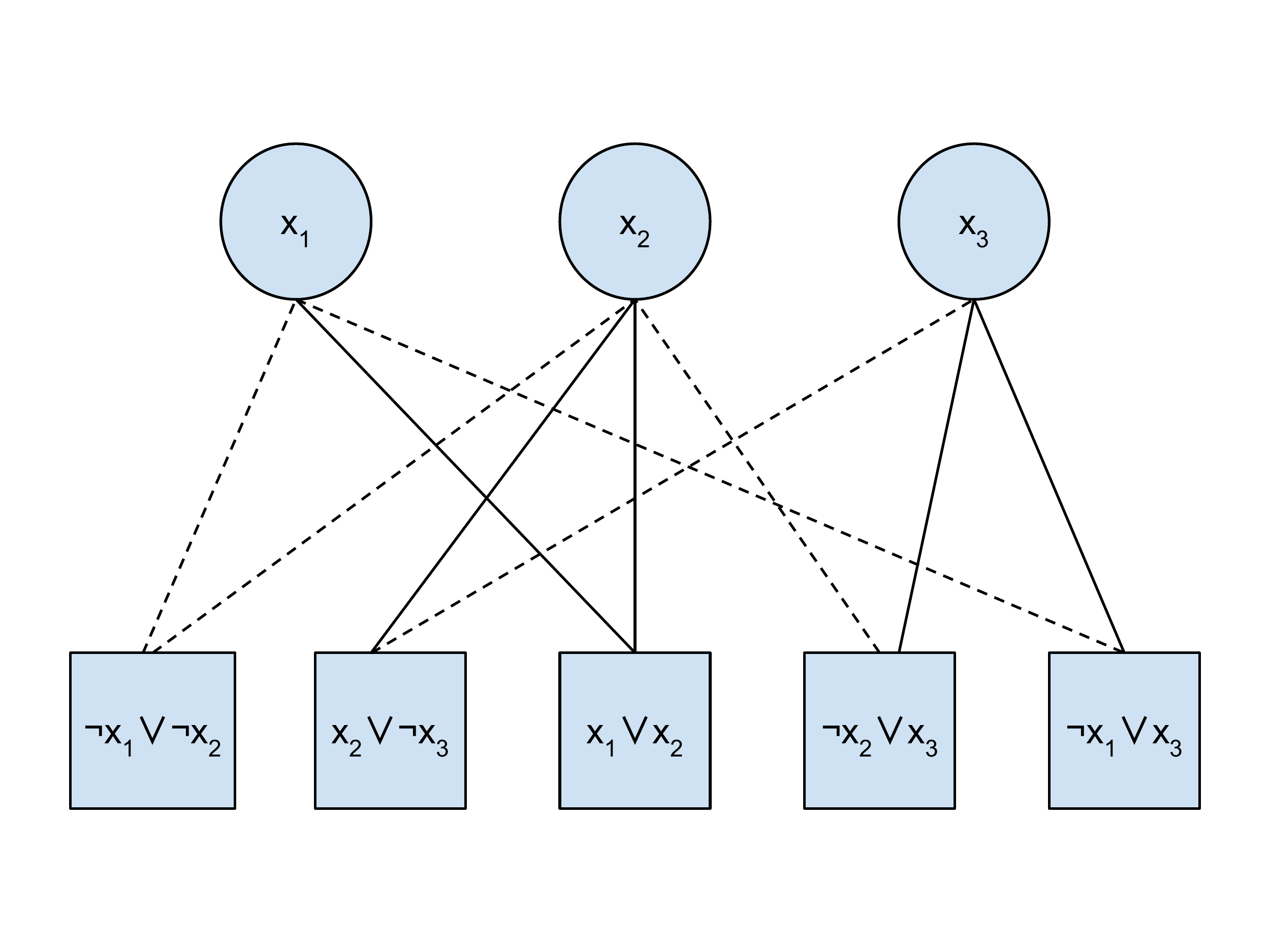}
	\caption{Factor-graph representation of CNF formula}
		\label{fig:factorgraph}
\end{figure}

There are at least two obvious representations of Boolean formulas as graphs. These are the familiar clause-variable factor graph (Figure \ref{fig:factorgraph}) and variable-variable graph. The former is an undirected bipartite graph that reserves one node type for clauses and another for nodes. In one instance of this setup, the edge label between a node and a clause denotes whether that node is negated or not in the clause. In the latter, nodes in a graph correspond exactly with the variables in a formula, and are connected by undirected edges if two nodes appear in at least one clause together. 
\subsection{CNF graphs and the phase shift}

To our knowledge there has not been any work that has studied, either directly or indirectly, the effect of the phase-shift control parameters on CNFs as graphs. 

 In one likely scenario, a weakly-supervised learner like a Graph Neural Net will be more effective at recognizing features of satisfiability in graphs describing CNF formulas that are distant from the phase shift, than for CNF formulas that are close to it. This effect is described by \citep{Devlin08b.:satisfiability}: Optimized search-solvers must perform more backtracking steps the closer they get to $\alpha_{cr}$. In another, more general scenario, the capacity to learn from graph representations will be different in some way at points before the phase shift, at the phase shift, and after it. 

\subsection{CNF graphs for GNNs}

From an implementation perspsective, the variable-variable type is the simpler of the two representations. In GNNs with multiple node types it is sensible to implement type-specific transition functions \citep{scarselli2009graph}. As we are just beginning to understand the behavior of GNNs we limit ourselves to a consideration of the variable-variable graph type in our experiments.

A variable-variable GNN graph $G$ contains twice the number of nodes in the Boolean formula $\phi$ with which it corresponds. Binary labels indicate whether a node is a negated literal or not ($x_{n}$ vs. $\neg x_{n}$), and equivalently indexed variables are connected by a special edge. Nodes $x_{n}$ and $x_{n'}$, for $n \neq n'$ are connected by an edge if they appear together in a clause in $\phi$. In our experiment, edges are given Euclidean labels $l_{n} \in \mathbbm{R}^{m}$ where $m$ is maximum number of clauses in the problems one chooses to consider. Entries $e_{i} = 1$ in an edge label $e$ if the two corresponding variables appear together in clause $i$, and equal $0$ otherwise.

\section{Data Generation and Experiments}

CNF formulas encode a very powerful and theoretically studied classifier: The clause-to-atom ratio. In order to determine whether GNNs can learn anything beyond this intrinsic classifier, we generate multiple training sets, each at a fixed clause-to-variable ratio. Additionally in order to analyze the effect of the phase shift in the learning ability of GNNs we chose to use uniformly randomly generated 3-SAT instances (RAND-SAT). These instances have a fixed number of clauses and atoms and exactly 3 literals per clause. Atoms appear in clauses with uniform probability and are negated with uniform probability.

We created 3 datasets with clause to atom ratios of (4.4,6.6,10). According to the clause-to-atom ratio, the probability that a formula is satisfiable is roughly 90\% in the first dataset, 50\% in the second and 10\% in the third. To prevent statistical skew from clouding the learning of features of satisfiability we created balanced datasets that were exactly 50\% satisfiable.

GNNs significantly improve over a random baseline for all three datasets. Interestingly, for the ratio closest to the phase-shift (where problems are thought to be the hardest) the network converged to a test accuracy of approximately 70\%, whereas the best accuracy for the two other problem sets was roughly 65\%. The exact errors are shown in Table \ref{tab:errs}. Figure 4 shows the training and validation error over time.

\begin{table}[h]
\centering

\begin{tabular}{llll}
Clause/Atoms & Train Error & Validation Error & Train Error \\
4.4          & 70.71\%     & 69.50\%          & 69.80\%     \\
6.6          & 65.65\%     & 66.30\%          & 64.40\%     \\
10           & 65.92\%     & 65.70\%          & 67.30\%
\end{tabular}
\caption{Training, Validation and Testing Error}
\label{tab:errs}
\end{table}

\begin{figure}
        \centering
        \begin{subfigure}[b]{0.3\textwidth}
                \includegraphics[width=\textwidth]{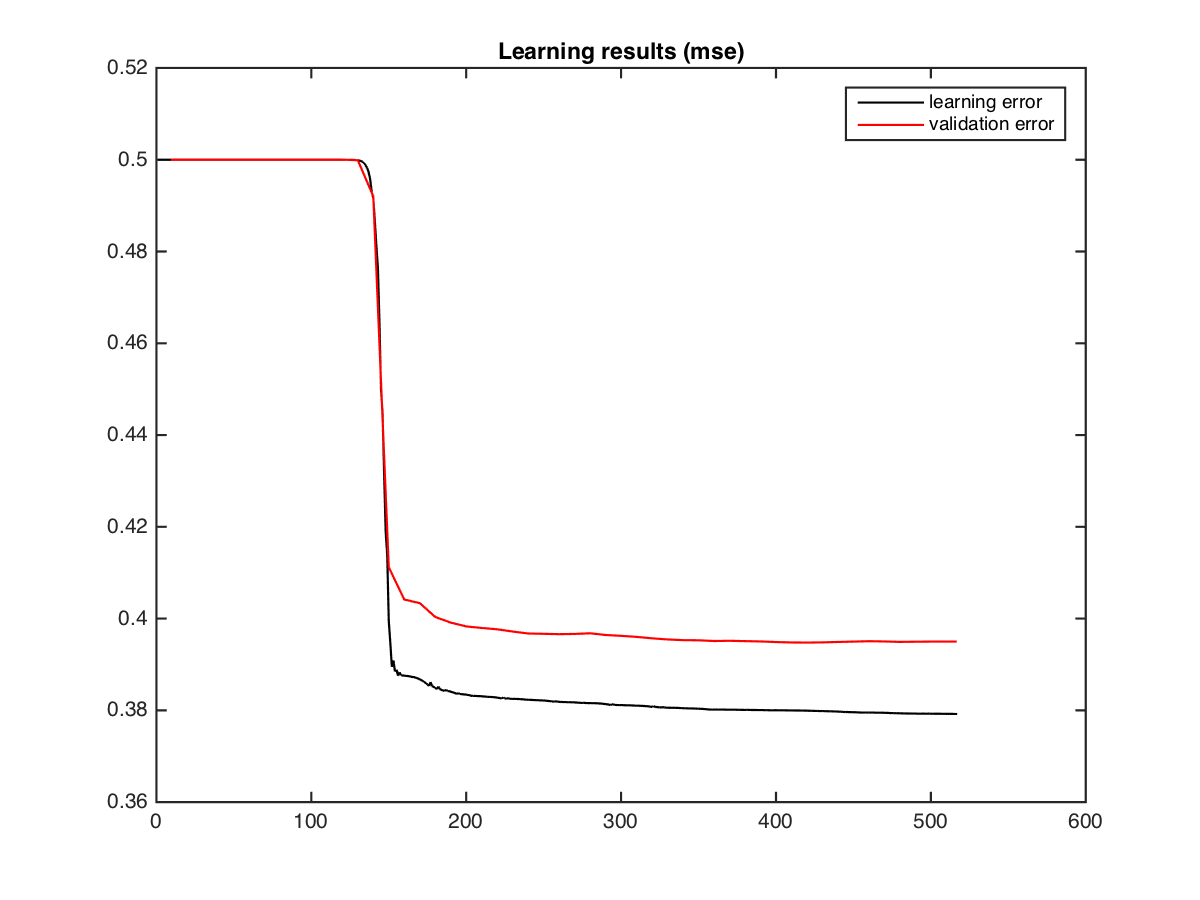}
                \caption{4.4 Clause Atom Ratio}
                \label{fig:1090learn}
        \end{subfigure}%
        ~ 
         \begin{subfigure}[b]{0.3\textwidth}
                \includegraphics[width=\textwidth]{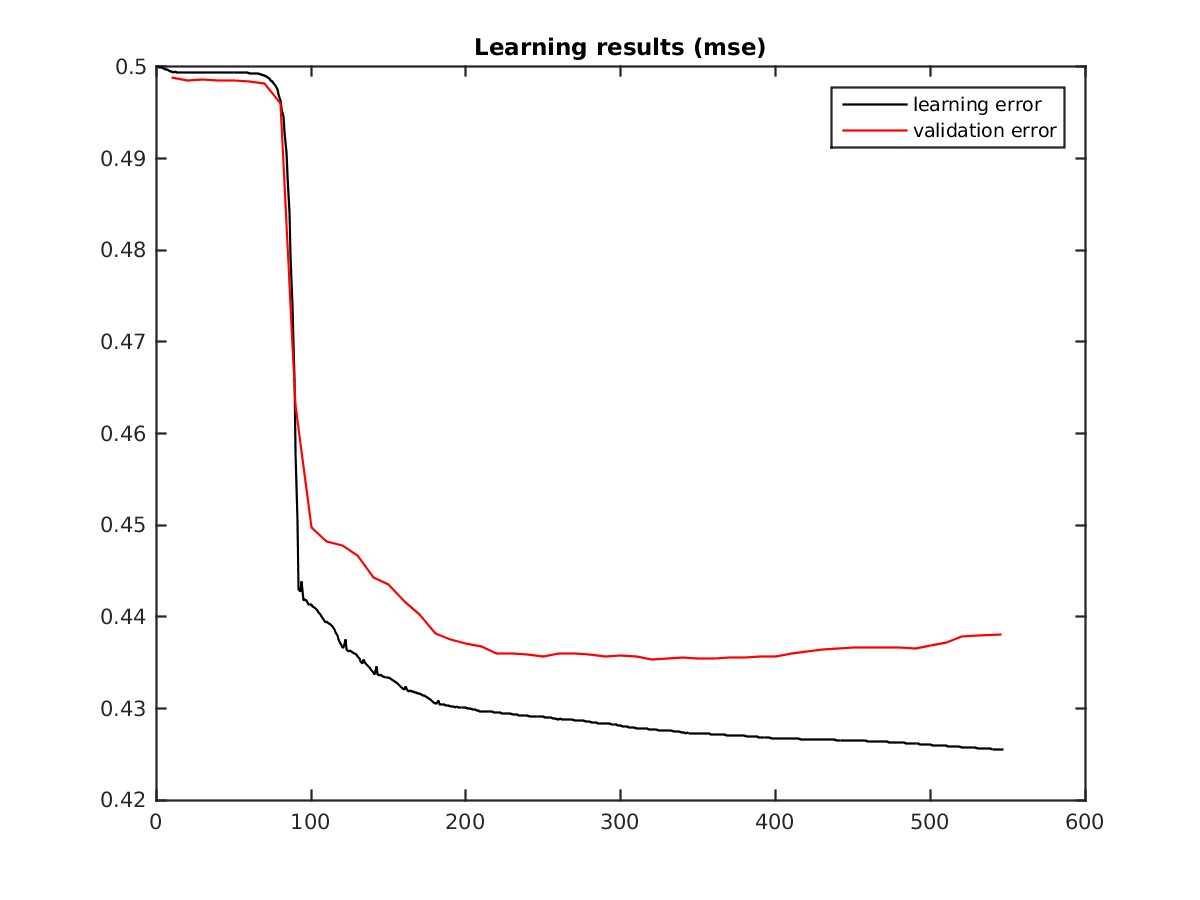}
                \caption{6.6 Clause Atom Ratio}
                \label{fig:5050learn}
        \end{subfigure}%
        ~ 
             \begin{subfigure}[b]{0.3\textwidth}
                \includegraphics[width=\textwidth]{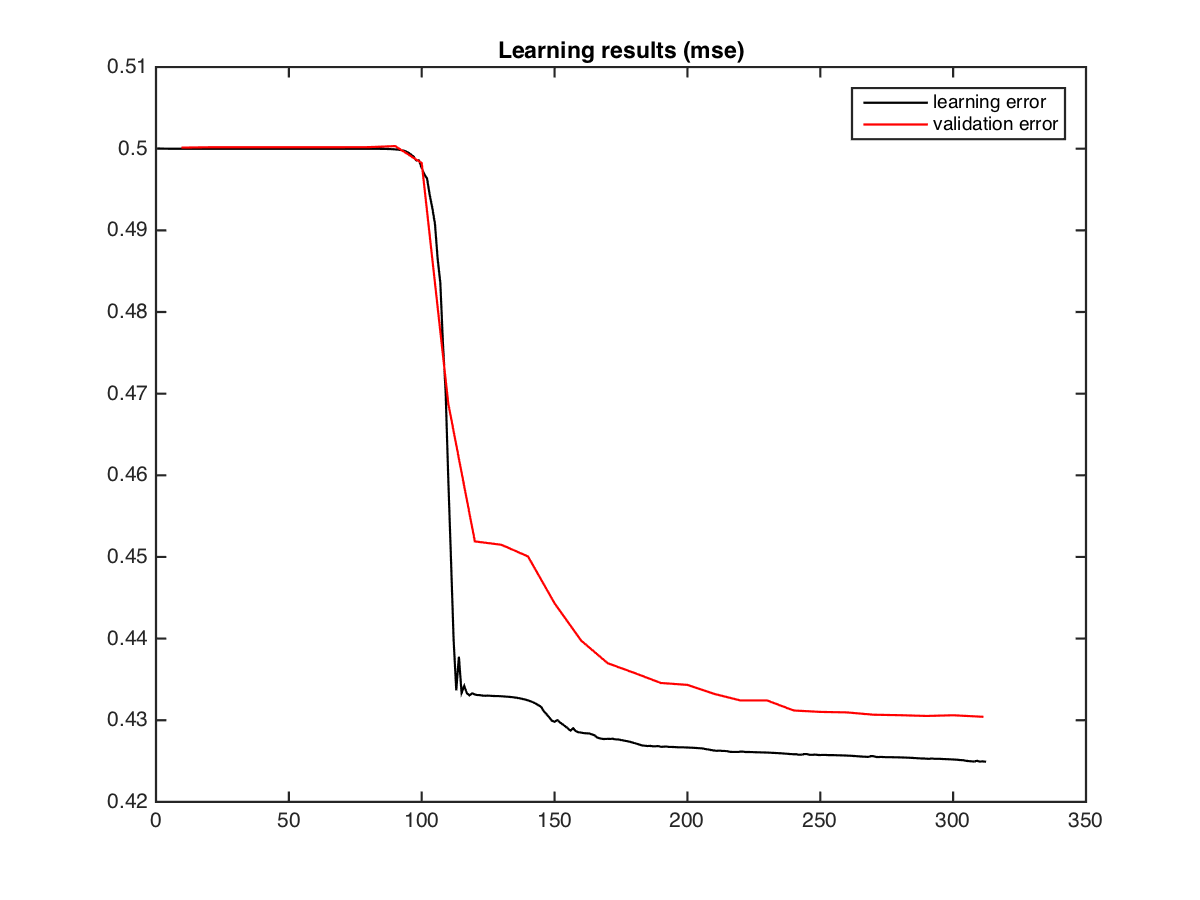}
                \caption{10 Clause Atom Ratio}
                \label{fig:9010learn}
        \end{subfigure}%
        \caption{Training and Validation error vs. epochs}\label{fig:learning}
\end{figure}

\section{Conclusions and Future Work}

In this paper we made nontrivial headway in applying deep learning with Boolean satisfiability. Initially, we sought to analogize Boolean satisfiability with natural language semantic classification, a domain in which neural networks have obtained state-of-the-art results for certain tasks. While our experiments to this effect were backgrounded in favor of later findings, they point to an interesting discovery: It does not seem that Boolean formulas in conjunctive normal form can, in any principled way, be made to ``look like" expressions in natural language. This discovery is coextensive with the finding that more general graph-like representations \textit{are} appropriate for representing the complex interdependencies necessary for classifying satisfiability.

We also apply Graph Neural Networks in a novel way. Our results suggest that graph classification may provide a new way of thinking about the theoretical properties of SAT, and that SAT can be used as a test domain for the expressive properties of neural learners. 

Fascinatingly, we find that without any explicit feature engineering Graph Neural Networks trained on a variable-variable representation of CNF formulas can in fact learn features of satisfiability, even for theoretically difficult problem instances.

There are several obvious directions for future research. As mentioned in our results section, we did not have the time to optimize for hyperparameters. For example, what kinds of nonlinearities are most appropriate for nonlinear, FNN transformation functions? The variable-variable graph as we define it is just one possible graph representation of CNF formulas. We intend to explore the effect of other representations on the ability of GNNs to recognize satisfiability.

\clearpage
\appendix
\section{Circuit Solvers}

As mentioned in Section 2 neural architectures can be used to represent combinatorial problems, such as SAT, as global optimization problems.

Consider the following SAT formulation. Let $x \in \{-1,1\}^N$ be a vector where the $j$'th element of $x$ represents Boolean variable $x_j$ and is thus $1$ if $x_i$ is true and $-1$ otherwise. A SAT instance can be described by a matrix $W$ where 
 $$W_{i,j}=\begin{cases}
 	1 & x_j \in \omega_i\\
 	-1 & \neg x_j \in \omega_i\\
 	0 & \text{otherwise}
 \end{cases}$$
Further consider the step activation function:
$$\theta(x)=\begin{cases}
	1 & \text{if } x\geq 0\\
	0 & \text{otherwise}
\end{cases} $$
Let $\textsc{sat}_W: \{-1,1\}^N\rightarrow \{0,1\}$, defined for a specific SAT instance $W$, be a function which maps an assignment vector $x$ to $1$ if $x$ satisfies the instance and $0$ otherwise. Specifically:
$$\textsc{sat}_W(x)=\theta(\sum_i\theta(W_{i,\cdot}\cdot x + W_{i,\cdot}^2\cdot x^2 - .5)-M+.5)$$
Note that
$$\max_x \textsc{sat}_W(x) \begin{cases}
	1 & \text{if } W \text{ is satisfiable}\\
	0 & \text{otherwise}
\end{cases}$$
 That is, the global optima of the following function correspond bijectively with satisfying assignments. More specifically $W_{i,\cdot}\cdot x+W_{i,\cdot}^2\cdot x^2-.5$ is greater than zero if and only if $W_{i,j}=x_{j}$ for some $j$. That is, if clause the disjunctive clause $i$ is satisfied. The objective value of $\textsc{sat}_W$ is 1 if and only if all clauses are satisfied.
 
 The idea of the circuit solver is to approximate the step activation function using $\sigma$\footnote{the sigmoid function} and additionally approximate the binary nature of $x$ by replacing $x$ with $tanh(\hat{x})$ where $\hat{x}\in \mathbb{R}^N$. Concretely, let:
 $$\textsc{approxsat}_W(\hat{x}):=\sigma(\sum_i \sigma(W_{i,\cdot}\cdot tanh(x)+W_{i,\cdot}^2\cdot tanh(x)^2-.5)-M+.5)$$
 $\textsc{approxsat}_W:R^N \rightarrow (0,1)$ retains some of the nice properties of $\textsc{sat}_W$ while being defined over $R^N$ and having well-defined gradients. Specifically, when interpreting $x_j$ as true if $x_j>0$ and false if $x_j<0$, note that $v_i\cdot tanh(x)+v_i^2\cdot tanh(x)^2-.5>0$ implies that $\text{sgn}(v_ij)=\text{sgn}(x_j)$ for some $j$ and thus that the conjunctive clause $i$ is satisfied. The objective value for a clause can only be positive if the clause is in fact satisfied. Additionally this implies that if $\textsc{approxsat}_W(\hat{x})>0 \implies \textsc{sat}_W(\text{sgn}(\hat{x}))=1$. The real valued $\hat{x}$ can be rounded to a satisfying assignment.\footnote{sgn is the signum function that returns the sign of a value}

 \subsection{Experiments}

\begin{figure}[htb]
 	\includegraphics[width=\textwidth]{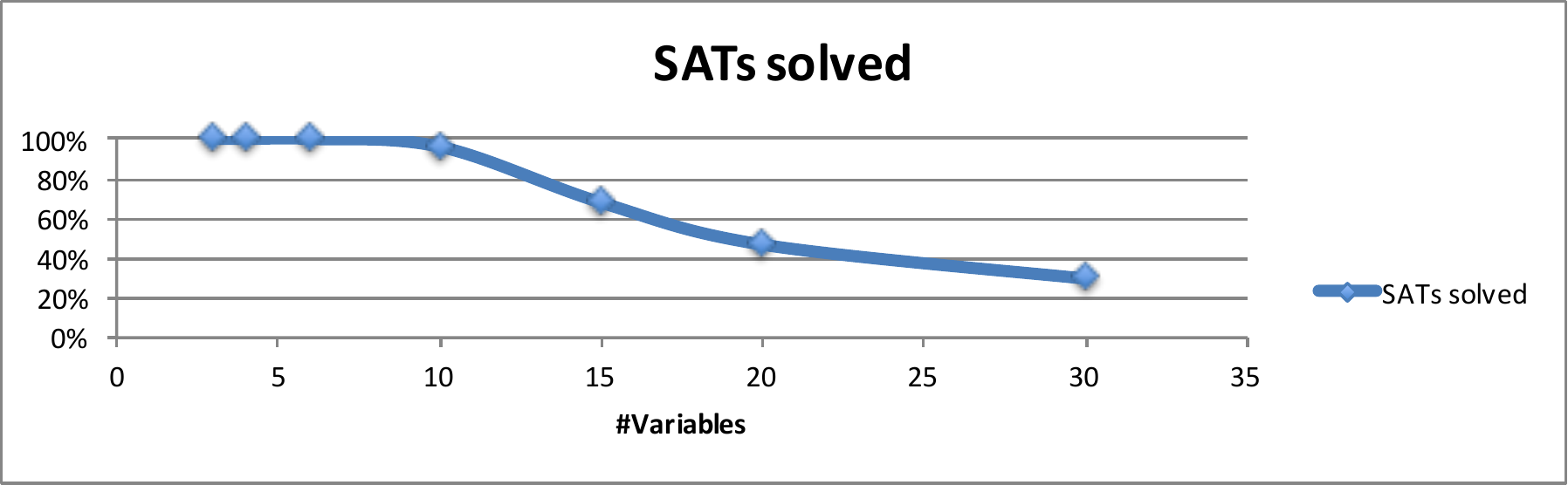}
\caption{Number of SAT's solved vs. number of variables. 100 instances per data point}
\label{fig:anneal}
 \end{figure}

 To test the neural sat encoding we ran a set of experiments. Random SAT problems are well known to be hard only for a very narrow ratio of clauses to variables. For 3-Sat that ratio is believed to be around 4.3 \citep{Saitta:2011:PTM:2024601} \citep{mlSAT}. To test the scalability of our optimization approach we tested how well we were able to find satisfying assignments, or determine unsatisfiability for different problem sizes. As stated before the procedure cannot result in false negatives, as an unsatisfying assignment could never be mistaken for a satisfying one. Figure \ref{fig:anneal} thus plots the false positive rate vs. different problem sizes. We can see that for small problem instances the annealing approach finds satisfying solutions for all instances that were in fact satisfiable. For larger instances, the approach begins to fail and is not successful.The optimization procedure gets stuck in local optima.
\section{Recursive Neural Networks}

As described in Section 5, we maintain that CNF formulas are best represented as graphs. Feed-forward neural networks are incapable of operating on graphs, as they are general function approximators for functions defined over Euclidean space. 

Recursive Neural Networks (RNNs) are capable of classifying tree-structures. As a first experiment we investigated whether the success of RNNs in natural language processing task had any implictions for classifiaction of Boolean formulas. In this setting we interpret disjunctive CNF clauses as words within a ``sentence," and a represent them using vector representations of clauses $\omega_{j}$defined in Appendix A.

Unlike sentences in natural language, Boolean formulas do not have a tree-like syntactic structure. The clauses in a formula are inherently commutable and their meaning composition is not recursive or scopal. In order to impose some tree-structure on Boolean formulas, we hierarchically cluster clause vectors based on their cosine similarity. Each node in the resulting binary tree contains a subset of the clauses of the original formula. The root then ``represents" the whole formula.

\begin{figure}[htb]
 	\centering
 	\includegraphics[scale=.5]{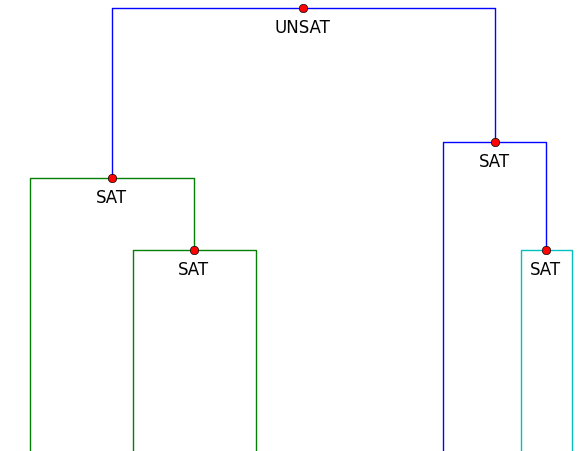}
\caption{Number of SAT's solved vs. number of variables. 100 instances per data point}
\label{fig:clust}
 \end{figure}

In keeping with the sentiment classification task of \cite{SocherEtAl2012:MVRNN}, each node in the tree is labeled according to the satisfiability of the subproblem. Note that if a node is labeled SAT then implicitly so must the whole subtree below it. An example is shown in Figure \ref{fig:clust}.

The RNN was not able to successfully learn any meaningful distinction between satisfiable and unsatisfiable formulas. As described in Section 3 the satisfiability is largely dependent on the clause to variable ratio. As each inner node of the RNN tree contains only a subset of the clauses and the same number of variables it is highly unlikely for such a subtree to be unsatisfiable. For this reason, even though the full problems were chosen to be balanced between SAT and UNSAT the cost function of the RNN was highly biased towards SAT. The RNN, thus, would guess every instance to be SAT and could not improve upon that.

We tested the setting in which only root nodes were labeled. Moreover, we tested other 
neural architectures with a somewhat tree-like, directed acyclic structure, 
but none of them showed any success over a random baseline.

\pagebreak
\bibliographystyle{aaai-named}
\bibliography{224d}
\end{document}